\documentclass[letterpaper]{article} 
\usepackage{aaai2026}  
\usepackage{times}  
\usepackage{helvet}  
\usepackage{courier}  
\usepackage[hyphens]{url}  
\usepackage{graphicx} 
\usepackage{natbib}  
\usepackage{caption} 
\usepackage{algorithm}
\usepackage{algorithmic}
\usepackage{amsmath}  
\usepackage{amssymb}  
\usepackage{multirow}
\usepackage{array}
\newcolumntype{M}[1]{>{\centering\arraybackslash}m{#1}}
\usepackage{makecell}
\usepackage{booktabs}
\usepackage{tabularx}
\usepackage{xspace} 
\newcolumntype{Y}{>{\centering\arraybackslash}X}  
\newcommand{\ours}{\texttt{FedCCCU}\xspace}
\nocopyright
\usepackage{times}
\usepackage{helvet}
\usepackage{courier}
\usepackage{xcolor}
\usepackage{newfloat}
\usepackage{listings}
\DeclareCaptionStyle{ruled}{labelfont=normalfont,labelsep=colon,strut=off} 
\floatstyle{ruled}
\newfloat{listing}{tb}{lst}{}
\floatname{listing}{Listing}
\title{Federated Unlearning in the Wild: Rethinking Fairness and Data Discrepancy}
\author{
    ZiHeng Huang\textsuperscript{\rm 1}\equalcontrib,
    Di Wu\textsuperscript{\rm 2}\equalcontrib,
    Jun Bai\textsuperscript{\rm 3}\thanks{Corresponding author: baijun@deakin.edu.au},
    Jiale Zhang\textsuperscript{\rm 4},
    Sicong Cao\textsuperscript{\rm 5},
    Ji Zhang\textsuperscript{\rm 6},
    Yingjie Hu\textsuperscript{\rm 1}
}
\affiliations {
    \textsuperscript{\rm 1} Central South University\\
    \textsuperscript{\rm 2} La Trobe University\\
    \textsuperscript{\rm 3} McGill University\\
    \textsuperscript{\rm 4} Yangzhou University\\
    \textsuperscript{\rm 5} Nanjing University of Posts and Telecommunications\\
    \textsuperscript{\rm 6} University of Southern Queensland
}

\usepackage{bibentry}

\begin{document}

\maketitle


\begin{abstract}

Machine unlearning has become a critical capability to support data deletion rights, such as the "right to be forgotten" mandated by privacy regulations. As a decentralized learning paradigm, Federated Learning (FL) also faces growing demands for unlearning. However, enabling unlearning in realistic FL settings presents two major challenges. 
First, fairness in FU is often overlooked: 1) Existing exact unlearning technical Routes typically require all clients to participate in retraining, regardless of their involvement in the unlearning request, leading to unnecessary computation and communication overhead; 2) recent approximate approaches apply gradient ascent, distillation or directly zero out neurons associated with the forget set, but such coarse interventions neglect their relevance to retained knowledge. This can unfairly degrade performance for clients whose data is entirely from the retain set.
Second, data distribution discrepancy poses a significant challenge. Most existing evaluations rely on artificially synthetic IID or non-IID assumptions that fail to reflect the natural heterogeneity of real-world federated systems. These unrealistic benchmarks obscure the true impact of unlearning on both local and global utility and limit the applicability of current methods in production environments. To bridge this gap, we conduct a comprehensive benchmark of existing FU technical Routes under both fairness and realistic data heterogeneity conditions. Furthermore, we propose a novel and fairness-aware FU approach, namely Federated Cross-Client-Constrains Unlearning (\ours), that explicitly addresses both challenges. \ours offers a practical and scalable solution for real-world FU, providing a foundation for future research in this area. Experimental results show that existing methods perform poorly under realistic data settings, while our approach consistently outperforms them across diverse, real-world scenarios.

\end{abstract}


\section{Introduction}


With the widespread adoption of machine learning, the need for user data deletion and compliance with privacy regulations has become increasingly critical. This demand is further reinforced by legal frameworks such as the GDPR~\cite{2} and CCPA~\cite{3}, both of which explicitly grant users the right to request data deletion. In response, Machine Unlearning (MU)~\cite{bourtoule2021machine, li2025machine} has emerged as a key technique for realizing the “right to be forgotten,” achieving promising results in centralized settings. Common strategies involve retraining or targeted model interventions to remove the influence of specific data from learned parameters. However, these centralized approaches are not directly applicable to Federated Learning (FL), where a global model is formed by aggregating locally trained updates from distributed clients. The collaborative and decentralized nature of FL fundamentally differs from the centralized paradigm, making retraining-based unlearning infeasible at the individual client level. Consequently, there is an urgent need to develop FU methods \cite{liu2024survey} that can ensure regulatory compliance while maintaining strong privacy guarantees in distributed environments.

Current federated unlearning research primarily follows two mainstream technical routes: exact unlearning and approximate unlearning. The exact unlearning route aims to restore the model to a state as if the target data had never been involved in training. This is typically achieved through strategies such as delete-and-retrain or label relabeling, which require global retraining involving all clients. Representative methods include FedEraser~\cite{4}, VeriFi~\cite{5}, SFU~\cite{6}, and KNOT~\cite{7}. These methods typically require all clients to participate in global retraining, resulting in substantial computational and coordination overhead. The approximate unlearning route seeks to balance unlearning effectiveness with practicality by avoiding full retraining. It encompasses a variety of techniques such as knowledge distillation, gradient editing, and model editing, which aim to suppress or erase the influence of specific data from the model. Representative methods in this category include MoDe~\cite{8}, GA~\cite{9,26}, FedFilter~\cite{10}, 2F2L~\cite{11}, FedRecovery~\cite{12}, and DEPN~\cite{13}. These approaches generally perform well in small-scale settings or under synthetically partitioned non-IID data, and currently represent the dominant direction of research in FU. 


However, existing paradigms rely on two fragile assumptions. First, current exact unlearning methods enforce global retraining across all clients regardless of their involvement, and approximate unlearning that overlooks the importance of retained knowledge, which together raise fairness concerns at both the system and model levels. Second, evaluations typically simulate non-IID conditions via label-based partitioning, which fails to reflect real-world cross-domain feature heterogeneity and thus overstates the robustness of current methods. 



\begin{figure}[htbp]
  \centering
  \includegraphics[width=0.9\linewidth]{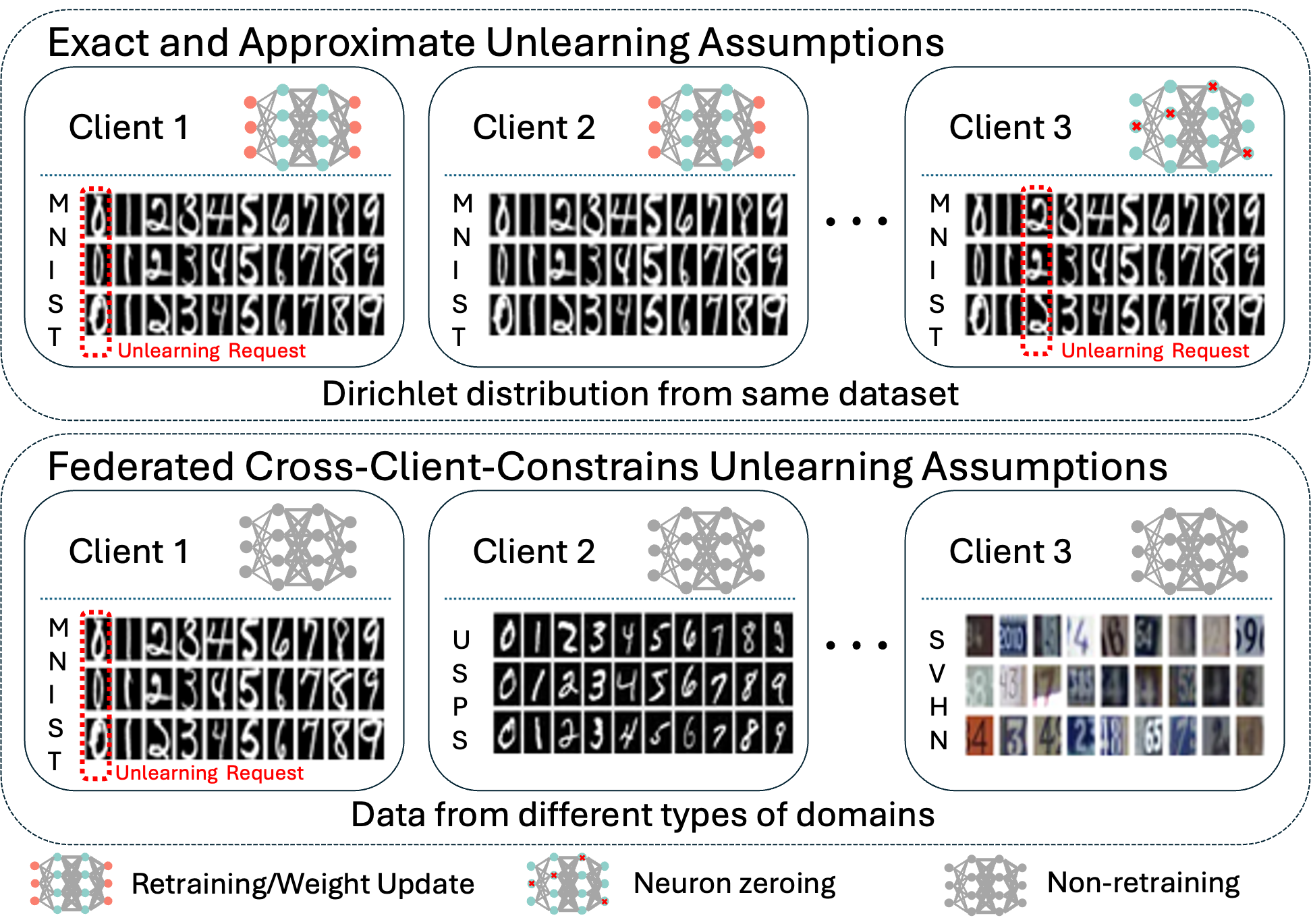}
  \caption{Federated unlearning in single-domain (top) and cross-domain (bottom) settings.}
  \label{fig1}
\end{figure}

Figure~\ref{fig1} illustrates the gap between conventional and realistic FU settings. Prior works often simulate non-IID data via Dirichlet-based label splits over a single dataset (“pseudo-Noniid”), ignoring real-world scenarios where clients from different domains (e.g., schools, banks, postal services) share label spaces but differ in feature distributions. Existing FU technical Routes typically enforce global retraining or model editing across all clients, assuming full cooperation and fairness, yet they either impose redundant costs on uninvolved clients or degrade unrelated knowledge through coarse neuron removal. In contrast, our Federated Cross-Client-Constrained Unlearning (\ours) addresses both issues by considering cross-domain heterogeneity and constraining model editing to minimize collateral impact on non-forgetting clients, making FU fairer and more practical. The contributions of this paper can be summarized below. 


\begin{itemize}
    \item We formalize two overlooked challenges in federated unlearning: fairness and data distribution discrepancy in the cross-domain settings for the different technical routes.
    \item We conduct an extensive evaluation of representative exact and approximate FU methods under our benchmark and reveal their limitations in both forgetting effectiveness and fairness in realistic federated learning scenarios.
    \item We propose a novel fairness-aware unlearning method (\ours) and advocate for future research to prioritize fairness and data realism alongside accuracy, paving the way for deployable and responsible federated unlearning systems.
\end{itemize}


\section{Related Work}
Existing FU research still follows a “retrain-all plus synthetic data” paradigm, and its evolution traces several intersecting lines. FedEraser\cite{4} cancels historical gradients by backtracking to replicate a full retrain, but the computational and costs are spread across every client, creating fairness issues. Subsequent work tried to lighten this load: Ferrari\cite{16} deletes sensitive features in embedding space; \cite{17} shard clients and encode within each shard to accelerate retraining; Deepobliviate\cite{24} quantizes residual memories and trims iterations online; \cite{25} use a first-order Taylor approximation of the loss to finish retraining in just a few rounds; and SIFU\cite{28} combines time rollback with sequential re-optimization, clipping rollback points via bounded sensitivity before incrementally updating the remaining data.

Although these strategies improve efficiency, they all assume that CIFAR10/100 “pseudo-noniid” slices adequately represent real, cross-island distributions. To reduce overhead further, researchers have proposed approximate unlearning techniques. Knowledge distillation and soft-label recovery (\cite{18}, MoDe\cite{8}, \cite{27}) fade memory by relaxing equivalence; gradient ascent and feature filtering (GA\cite{9}\cite{16}, FedFilter\cite{10}) directly intervened in gradient directions; pruning methods (RevFRF\cite{29}, \cite{30}) excise class-specific weights and then fine-tune; FedRecovery\cite{12} injected differential privacy noise into residual gradients, while Depn\cite{13} provided millisecond-level responses through targeted weight pruning. VeriFi\cite{5} augments participation-heavy frameworks with zero-knowledge proofs, enhancing auditability but further increasing the resource burden imposed by global retraining. In sum, current FU techniques perform well on small, homogeneous datasets but remain untested in truly heterogeneous, real-world environments.

In contrast to unlearning tasks, the mainstream FL community has extensively explored fairness in resource allocation and performance, exemplified by personalized alignment approaches (e.g., FedDyn\cite{19}, FedBN\cite{20}). Concurrently, robust optimization methods addressing data heterogeneity (Karimireddy\cite{21}) and realistic non-iid benchmarks (such as Leaf\cite{22} and CORA\cite{23}) have become increasingly mature. However, these advances have not yet been transferred to the unlearning domain.

The abovementioned FU methods lack comprehensive consideration of realistic data distributions and unlearning deployment patterns, as summarized in Table~\ref{tab:assumptions}. Most existing FU approaches are still confined to synthetic IID or pseudo-noniid splits, and their retraining strategies fall into only two categories: (i) full-scale retraining that involves every client, or (ii) partial retraining in which, beyond the forgetting client, a subset of additional clients also participates.


\begin{table}[]
\centering
\footnotesize 
\setlength{\tabcolsep}{4pt} 
\renewcommand{\arraystretch}{1.1}
\begin{tabular}{lccl}
\hline
\multicolumn{1}{c}{\textbf{Method}} & \textbf{Year} & \textbf{\makecell{Multi-client \\ Retraining}} & \multicolumn{1}{c}{\textbf{\makecell{Data \\ Distribution}}} \\ \hline
FedEraser & 2021 & $\sqrt{}$ & IID / pseudo-noniid \\
Deepobliviate & 2021 & $\sqrt{}$ & pseudo-noniid \\
Revfrf & 2021 & $\sqrt{}$ & pseudo-noniid \\
Wu et al & 2022 & $\sqrt{}$ & pseudo-noniid \\
Liu et al & 2022 & $\sqrt{}$ & pseudo-noniid \\
Wang et al & 2022 & $\sqrt{}$ & pseudo-noniid \\
MoDe & 2023 & $\times$ & pseudo-noniid \\
FedFilter & 2023 & $\sqrt{}$ & pseudo-noniid \\
FedRecovery & 2023 & $\times$ & pseudo-noniid \\
KNOT & 2023 & $\sqrt{}$ & IID / pseudo-noniid \\
FedLU & 2023 & $\sqrt{}$ & pseudo-noniid \\
Ferrari & 2024 & $\times$ & IID \\
Lin et al & 2024 & $\sqrt{}$ & IID / pseudo-noniid \\
VeriFi & 2024 & $\times$ & pseudo-noniid \\
SIFU & 2024 & $\sqrt{}$ & pseudo-noniid \\
\textbf{\ours(Ours)} & – & $\times$ & \textbf{real-noniid} \\ \hline
\end{tabular}
\caption{The assumptions of different FU methods. ``Multi-client Retraining'' indicates whether retraining involves clients beyond the one requesting unlearning. ``Pseudo-noniid'' splits a single dataset across clients, while ``Real-noniid'' assigns distinct datasets to distinct clients.}
\label{tab:assumptions}
\end{table}

\section{Problem Formulation}

\subsection{Technical Routes}

\begin{itemize}
    \item \textbf{Delete-Retrain}: Considered the “gold standard” for exact unlearning, this route removes all sensitive samples from the forgetting client's local dataset and retrains the model globally. 
    \item \textbf{Relabel-Poison}: this route avoids deleting data. Instead, it randomly rewrites the labels of class-0 samples to other non-target classes, gradually weakening the model’s original decision boundary. 
    \item \textbf{Neuron-Zeroing}: this  route first performs a sensitivity analysis on the global model to identify the most activated neurons or channels for class-0 during forward propagation. These parameters are then zeroed out.
\end{itemize}

\subsection{Fair Retraining Dilemma}
In a typical federated learning system, a central server collaborates with K clients to minimize the global risk, formally defined as:

\begin{equation}
F(\mathbf{w}) = \sum_{k=1}^{K} \frac{n_k}{n} \, \mathbb{E}_{(x, y) \sim \mathcal{D}_k} \left[ \ell(\mathbf{w}; \mathbf{x}, y) \right]
\end{equation}


where $\mathbf{w}$ denotes the model parameters, $D_k$ represents the local dataset of client $k$ containing $n_k$ samples, and $n = \sum_{k=1}^{K} n_k$. The global loss is computed as the weighted aggregation of each client’s expected loss $E_k$, proportional to their local data sizes. Training proceeds until the validation error falls below a predefined threshold $\varepsilon$, and the communication round at which this first occurs is recorded as the convergence round $T_0$.

If a subset of clients $C_{\text{req}}$ later requests data removal, existing methods require retraining across all clients. This imposes unnecessary computational and communication overhead on non-requesting clients, raising fairness concerns. 


\subsection{The Pseudo-Noniid Fallacy}
Existing studies typically partition a single dataset $D$ among multiple clients via Dirichlet sampling. Although this approach modifies each client's label priors $p_k(y)$, it implicitly assumes all clients share an identical conditional distribution. However,  Real-Noniid scenario often differ significantly: even when tasks and labels uniformly involve , data distributions across clients, such as chalkboard photos, touchscreen signatures, and scanned images, vary substantially due to differing imaging processes. It means $p_k(x \mid y) \ne p_j(x \mid y)$.

\section{Cross-Domain FU Benchmark Across Technical Routes}


To benchmark the performance of the existing technical route on cross-domain FU. We propose Cross Domain FU, a benchmarking framework designed for realistic federated unlearning. Unlike prior settings that partition a single dataset, it treats each client as an autonomous data silo with heterogeneous features and labels. Upon an unlearning request, only the requesting clients retrain locally, while others retain previous model states. The server then aggregates updates until convergence. This protocol, based on local retraining and global inheritance, better reflects real-world deployment by aligning both data distribution and training dynamics with practical constraints.



\subsection{Dataset Overview and Selection Rationale}

In constructing the Cross Domain FU Benchmark, we adhere to a core principle: ensuring that all clients perform an identical classification task while maximizing divergence in their local feature distributions.

We select six widely used datasets, namely MNIST10 \cite{mnist}, SVHN \cite{svhn}, USPS \cite{usps}, CIFAR10 \cite{cifar}, CIFAR100 \cite{cifar}, and ImageNet \cite{imagenet}, whose basic statistics are summarized in Table~\ref{tab:datasets}. To ensure label consistency across domains, we retain only the overlapping classes, resulting in 9 shared labels between CIFAR10 and ImageNet, and 65 between CIFAR100 and ImageNet. These dataset pairs maintain a unified label space while presenting significant visual disparities. For instance, SVHN contains cluttered RGB backgrounds, whereas MNIST consists of clean binary images; similarly, ImageNet features rich textures, in contrast to the compressed visual patterns of CIFAR. This cross-domain variation offers a realistic foundation for evaluating federated unlearning under heterogeneous feature distributions.

\begin{table}[]
\centering
\footnotesize
\renewcommand{\arraystretch}{1.2}
\setlength{\tabcolsep}{4pt} 
\begin{tabular}{ccccc}
\hline
\textbf{Task} & \textbf{Dataset} & \textbf{\makecell{Number of\\Classes}} & \textbf{\makecell{Training\\Samples}} & \textbf{\makecell{Test\\Samples}} \\ \hline
\multirow{3}{*}{\makecell{Handwritten\\Digit\\Recognition}} & MNIST10 & 10 & 60000 & 10000 \\
 & SVHN & 10 & 73257 & 26032 \\
 & USPS & 10 & 7291 & 2007 \\ \hline
\multirow{3}{*}{\begin{tabular}[c]{@{}c@{}}Image\\ Recognition\end{tabular}} & CIFAR10 & 10 & 50000 & 10000 \\
 & CIFAR100 & 100 & 50000 & 10000 \\
 & ImageNet & 1000 & 1280000 & 50000 \\ \hline
\end{tabular}%
\caption{Overview of the Dataset}
\label{tab:datasets}
\end{table}

\subsection{Data distribution strategy}

To better address the “data realism gap,” we design a Real-Noniid partitioning strategy that maintains task consistency while inducing realistic feature heterogeneity across clients. Client-wise data allocations under each strategy are detailed in Table~\ref{tab:client_data_final}.

In the  task, we assign MNIST10, SVHN, and USPS to clients C1–C3, C4–C6, and C7–C9 respectively, with intra-group samples partitioned via a Dirichlet distribution. Each client thus receives data from a distinct source domain.

For image recognition, we construct two intersection-based scenarios: one with 9 shared classes from CIFAR10 and ImageNet, and another with 65 shared classes from CIFAR100 and ImageNet. In both cases, we allocate low-resolution CIFAR images (32×32) to five clients and high-resolution ImageNet images (224×224) to the other five.

\begin{table}[htbp]
\centering
\footnotesize
\renewcommand{\arraystretch}{1.2} 
\setlength{\tabcolsep}{6pt} 

\begin{tabular}{@{}llll@{}} 
\toprule
\textbf{\makecell{Split\\Scenario}} & \textbf{Client} & \textbf{\makecell{Dataset /\\Range}} & \textbf{Resolution} \\ 
\midrule
\multirow{7}{*}{Real-Noniid} & C1-C3 & MNIST10(0–9) & 28×28 \\
& C3-C6 & SVHN(0–9) & 32×32 \\
& C7-C9 & USPS(0–9) & 16×16 \\ 
\cmidrule(l){2-4} 
& C1-C5 & CIFAR10(0–8) & 32×32 \\
& C6-C10 & ImageNet(0-8) & 224×224 \\ 
\cmidrule(l){2-4}
& C1-C5 & CIFAR100(0–64) & 32×32 \\
& C6-C10 & ImageNet(0-64) & 224×224 \\ 
\bottomrule
\end{tabular}
\caption{Client-wise data allocation results under different split scenarios}
\label{tab:client_data_final}
\end{table}

This 'task-aligned but domain-divergent' setup ensures that Real-Noniid maintains label comparability while introducing multi-dimensional distribution shifts across factors such as resolution, illumination, texture, and capture modality, which are frequently encountered in real-world cross-organization federated systems.

\begin{figure}[htbp]
  \centering
  \includegraphics[width=0.9\linewidth]{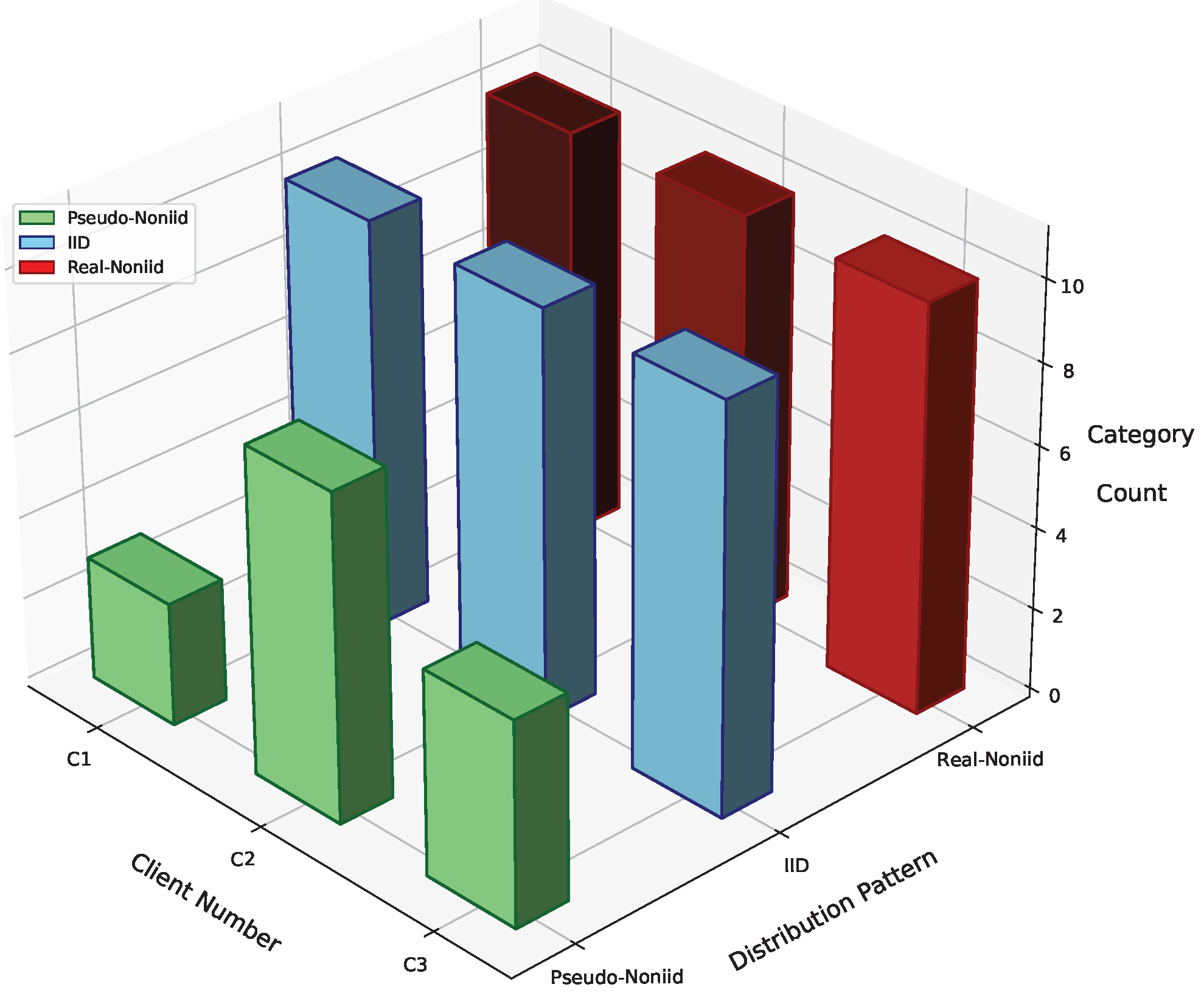}
  \caption{The distribution of data feature characteristics of the client in different allocation scenarios}
  \label{fig3}
\end{figure}

To visualize the differences between our Real-Noniid partition and existing data distribution assumptions, Figure~\ref{fig3} illustrates a 3D bar chart, where the x-axis denotes client IDs, the y-axis indicates the number of classes per client, and the z-axis represents average image resolution as a proxy for feature heterogeneity. In the IID setting, all bars are uniform in both height and depth. In Pseudo-Noniid, class distributions vary across clients, but resolution remains constant. In contrast, Real-Noniid maintains equal label coverage while exhibiting significant differences in resolution, highlighting the most realistic and challenging distribution scenario.

\begin{table*}[htbp]
\centering
\renewcommand{\arraystretch}{1}
\setlength{\tabcolsep}{4pt} 
\footnotesize
\begin{tabularx}{\textwidth}{@{}llll *{8}{>{\centering\arraybackslash}X}@{}}
\toprule
\multicolumn{12}{c}{\textbf{Evaluation Results under Real-Noniid Setting}} \\
\midrule
\multirow{3}{*}{\textbf{Data1}} & \multirow{3}{*}{\textbf{Data2}} & \multirow{3}{*}{\textbf{Data3}} & \multirow{3}{*}{\textbf{Model}} & \multicolumn{4}{c}{\textbf{\makecell{Global Accuracy}}} & \multicolumn{4}{c}{\textbf{\makecell{Client C1 (Class-0) Accuracy}}} \\
\cmidrule(lr){5-8} \cmidrule(lr){9-12}
& & & & \textbf{Original} & \textbf{Delete} & \textbf{Relabel} & \textbf{Zeroing} & \textbf{Original} & \textbf{Delete} & \textbf{Relabel} & \textbf{Zeroing} \\
\midrule
MNIST10 & SVHN & USPS & CNN & 94.09\% & 95.70\% & 95.59\% & 87.16\% & 98.95\% & 97.90\% & 96.50\% & 0.00\% \\
CIFAR10 & ImageNet & -- & ResNet18 & 90.88\% & 88.46\% & 88.62\% & 73.66\% & 94.33\% & 82.77\% & 83.67\% & 7.94\% \\
-- & ImageNet & CIFAR100 & ResNet32 & 69.46\% & 72.99\% & 69.90\% & 60.31\% & 62.86\% & 45.71\% & 51.43\% & 0\% \\ 
\bottomrule
\end{tabularx}

\caption{Evaluation results under the Real-Noniid setting. Only the client issuing the forgetting request is retrained; all others retain their pretrained global model. Delete-Retrain, Relabel-Poison, and Neuron-Zeroing represent three unlearning strategies.}
\label{tab:realnoniid_results}
\end{table*}

\subsection{Benchmark Experiments}


Previous studies have confirmed the effectiveness of the existing federal forgetting learning techniques in IID and pseudo-Noniid scenarios, Table~\ref{tab:realnoniid_results} reports results under the \textit{Real-Noniid} setting. Under the Real-Noniid setting, mainstream federated unlearning technical routes, including exact retraining and approximate model editing, show clear limitations. In , exact retraining fails to achieve effective forgetting, with target class accuracy remaining above 96\%, while approximate editing reduces it to 0\% but causes a 7\% drop in overall accuracy and degrades the model's accuracy on remaining classes.
In image recognition, exact retraining yields only partial forgetting (a 10–17\% drop) with performance loss, while approximate editing is more aggressive (e.g., from 94.33\% to 7.94\% or from 60.31\% to 0\%) but severely harms other classes and clients.

Table~\ref{tab:client_digit_rec_accuracy} and Table~\ref{tab:client_image_rec_accuracy} shows the accuracy variations of different clients on label\_0 data before and after applying federated unlearning in various tasks. By comparing the effects of different unlearning strategies, we observe the following key points.

In the  task, mainstream precise unlearning techniques failed to significantly reduce label\_0 accuracy on the forgotten clients due to high inter-client data homogeneity. In contrast, approximate techniques such as Neuron-Zeroing reduced the accuracy to 0\% across all clients, including those not involved in the unlearning request, reflecting a typical case of over-forgetting.
In the image recognition task, both mainstream unlearning technical routes led to limited forgetting on the target client. Specifically, on client0, label\_0 accuracy only decreased from 94.33\% to 82.77\% and from 62.86\% to 45.71\%, indicating a partially effective but insufficient forgetting outcome. Meanwhile, performance on non-target clients also deteriorated, with label\_0 accuracy on client9 dropping by 26\%. This suggests that local unlearning perturbations may propagate through model aggregation, resulting in unfair cross-client degradation.

\begin{table}[htbp]
\centering
\renewcommand{\arraystretch}{0.9}
\footnotesize
\setlength{\tabcolsep}{3pt}
\begin{tabularx}{\columnwidth}{@{}ll *{4}{>{\raggedleft\arraybackslash}X}@{}}
\toprule
\multicolumn{1}{c}{\textbf{Task}} & \multicolumn{1}{c}{\textbf{Client}} & \textbf{\makecell{Original}} & \textbf{Delete} & \textbf{Relabel} & \textbf{Zeroing} \\ 
\midrule
\multirow{9}{*}{\makecell{Handwriting\\Digit\\Recognition}} & client1 & 98.77\% & 97.90\% & 96.50\% & 0.00\% \\
& client2 & 99.36\% & 98.08\% & 95.83\% & 0.00\% \\
& client3 & 98.97\% & 97.94\% & 98.97\% & 0.00\% \\ 
\cmidrule(l){2-6} 
& client4 & 96.35\% & 97.08\% & 96.35\% & 0.00\% \\
& client5 & 95.72\% & 96.20\% & 94.06\% & 0.00\% \\
& client6 & 96.99\% & 96.99\% & 94.64\% & 0.00\% \\ 
\cmidrule(l){2-6} 
& client7 & 28.57\% & 53.57\% & 89.29\% & 0.00\% \\
& client8 & 26.13\% & 52.26\% & 90.24\% & 0.00\% \\
& client9 & 27.27\% & 52.27\% & 90.91\% & 0.00\% \\ 
\bottomrule
\end{tabularx}
\caption{Per-client Label\_0 accuracy for the Handwriting Digit Recognition task before and after unlearning.}
\label{tab:client_digit_rec_accuracy}
\end{table}

Nevertheless, in some specific cases, unlearning strategies may have positive effects. For example, in the Image Recognition(65) task, the accuracy of client5 and client8 improved from 60\% and 33.33\% to 73.33\% and 66.67\%, respectively. This phenomenon indicates that the unlearning process not only removes irrelevant or conflicting gradients but can also serve as a regularization technique, improving model performance on heterogeneous clients. 

In summary, existing FU strategies exhibit significant variability in their effects in real-world applications, particularly when data distribution heterogeneity is high. The performance of precise and approximate unlearning strategies across different tasks reveals both positive and negative effects that may arise when models face multi-client data heterogeneity.

\begin{table}[htbp]
\centering
\renewcommand{\arraystretch}{0.9}
\footnotesize
\setlength{\tabcolsep}{3pt}
\begin{tabularx}{\columnwidth}{@{}ll *{4}{>{\raggedleft\arraybackslash}X}@{}}
\toprule
\multicolumn{1}{c}{\textbf{Task}} & \multicolumn{1}{c}{\textbf{Client}} & \textbf{\makecell{Original}} & \textbf{Delete} & \textbf{Relabel} & \textbf{Zeroing} \\ 
\midrule
\multirow{10}{*}{\makecell{Image\\Recognition\\(9)}} & client1 & 94.33\% & 82.77\% & 83.67\% & 7.94\% \\
& client2 & 100.00\% & 77.78\% & 77.78\% & 5.56\% \\
& client3 & 94.05\% & 83.78\% & 84.86\% & 9.73\% \\
& client4 & 94.59\% & 83.78\% & 85.41\% & 8.65\% \\
& client5 & 97.08\% & 85.96\% & 83.04\% & 8.19\% \\ 
\cmidrule(l){2-6} 
& client6 & 90.32\% & 83.87\% & 80.06\% & 17.01\% \\
& client7 & 92.41\% & 79.11\% & 83.54\% & 17.72\% \\
& client8 & 90.74\% & 77.78\% & 85.19\% & 14.81\% \\
& client9 & 100.00\% & 73.91\% & 73.91\% & 17.39\% \\
& client10 & 93.55\% & 79.03\% & 84.68\% & 16.94\% \\ 
\midrule
\multirow{10}{*}{\makecell{Image\\Recognition\\(65)}} & client1 & 62.86\% & 45.71\% & 51.43\% & 0.00\% \\
& client2 & 57.14\% & 57.14\% & 71.43\% & 0.00\% \\
& client3 & 50.00\% & 50.00\% & 50.00\% & 0.00\% \\
& client4 & 53.85\% & 38.46\% & 53.85\% & 0.00\% \\
& client5 & 76.74\% & 62.79\% & 69.77\% & 0.00\% \\ 
\cmidrule(l){2-6}
& client6 & 60.00\% & 73.33\% & 73.33\% & 0.00\% \\
& client7 & 75.00\% & 50.00\% & 75.00\% & 0.00\% \\
& client8 & 80.70\% & 78.95\% & 78.95\% & 0.00\% \\
& client9 & 33.33\% & 66.67\% & 66.67\% & 0.00\% \\
& client10 & 71.43\% & 61.90\% & 61.90\% & 0.00\% \\ 
\bottomrule
\end{tabularx}
\caption{Per-client Label\_0 accuracy for Image Recognition tasks before and after unlearning. Image Recognition(9/65) denotes tasks with 9 and 65 classes.}
\label{tab:client_image_rec_accuracy}
\end{table}

\section{Proposed Method}
\begin{figure}[htbp]
  \centering
  \includegraphics[width=0.9\linewidth]{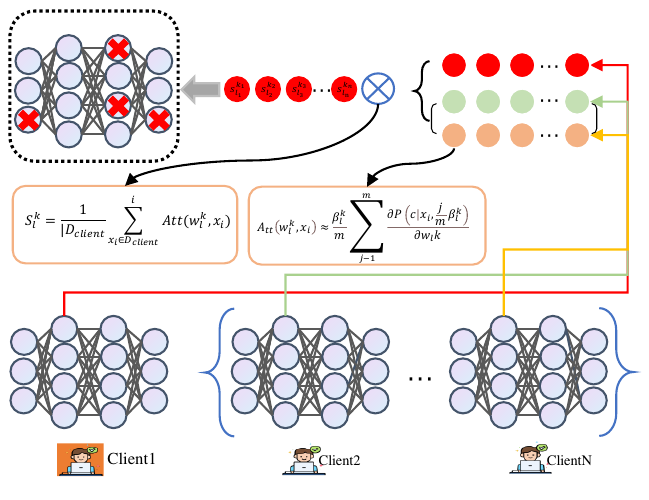}
  \caption{FedCCCU: Federated Cross-Client-Constrains Unlearning}
  \label{fig:method}
\end{figure}

To address the critical deficiencies of existing FU methods in terms of fairness and the modeling of data distributions, we propose a novel FU method named Federated Cross-Client-Constrains Unlearning(\ours). This method is specifically designed for realistic cross-domain data distribution scenarios, aiming to achieve effective unlearning while minimizing the performance impact on non-requesting clients. \ours introduces a cross-client constraint mechanism, which, combined with a lightweight model editing strategy, enhances the method's deployability and robustness. Figure~\ref{fig:method} illustrates the overall workflow of \ours.

\subsection{Identification of Key Neurons}
Our method for identifying key neurons associated with specific classes is primarily inspired by the DEPN\cite{13} framework. DEPN effectively quantifies the contribution of individual neurons to the model’s output through a gradient-based attribution approach. We apply this concept within our federated learning framework, aiming to identify sensitive neurons in the model that are highly associated with the class that needs to be forgotten (e.g., label=0).

The entire identification process is executed locally at each client within the federated learning framework, ensuring the privacy of client data is preserved. The specific procedure is as follows:  
When the server receives a forgetting request from a client, each client will receive the current global model parameters, denoted as $\theta$. The client’s objective is to compute the sensitivity score for each neuron in the model, for every class, based on the data set $D_{client}$ it holds locally.

For any neuron $w$ in the model (where $l$ represents the layer index and $k$ is the neuron index within that layer), its contribution to classifying a single data sample $x_i$ as the target class $c$ can be quantified by calculating the cumulative gradient of the class prediction probability $P(c|x_i, \theta)$ as its activation value changes from 0 to its original value $\beta$. This contribution, referred to as the Attribution Score, is computed as follows:

\begin{equation}
\operatorname{Att}(w_l^k, x_i) = \beta_l^k \int_{0}^{\beta_l^k} \frac{\partial P(c|x_i, \alpha_l^k)}{\partial w_l^k} d\alpha_l^k
\end{equation}

\begin{itemize}
    \item $\beta_i^k$ is the original activation value of neuron $w_i^k$ when the input is $x_i$.
    \item $P(c|x_i, \alpha_i^k)$ represents the probability that the model predicts input $x_i$ as class $c$ when the activation value of neuron $w_i^k$ is temporarily set to $\alpha_i^k$.
    \item $\frac{\partial P(\cdot)}{\partial w_i^k}$ is the partial derivative of the class prediction probability with respect to the neuron $w_i^k$, i.e., the gradient.
\end{itemize}

Given that directly computing the continuous integral is difficult, we approximate it using the Riemann sum, discretizing the integration process into $m$ steps:

\begin{equation}
\operatorname{Att}(w_l^k, x_i) \approx \frac{\beta_l^k}{m} \sum_{j=1}^{m} \frac{\partial P(c|x_i, \frac{j}{m}\beta_l^k)}{\partial w_l^k}
\end{equation}

\begin{itemize}
    \item $\frac{j}{m}\beta_l^k$ represents the activation value of the neuron at the $j$-th discrete step.
\end{itemize}

The client calculates attribution scores for all neurons for each sample in its local dataset $D_{client}$. Subsequently, by averaging the attribution scores of all samples in the dataset, a final sensitivity score $S_l^{k}$ is obtained for each neuron $w_l^k$ on that client.

\begin{equation}
S_l^k = \frac{1}{\left| D_{client} \right|} \sum_{x_i \in D_{client}} \operatorname{Att}(w_l^k, x_i)
\end{equation}

\begin{itemize}
    \item $|D_{client}|$ is the total number of samples in the client's local dataset $D_c^{client}$.
\end{itemize}

For a given class $c$, a neuron’s score indicates the strength of its association with that class, with higher scores reflecting stronger associations. In the end, each client will upload the indices (l, k) of the topN most sensitive neurons, along with their corresponding sensitivity scores $S$, for each class to the central server.

\subsection{Neuron Dominant Computing}
In this section, we propose a method to measure the importance of neurons in the forgetting and non-forgetting clients. Based on this new idea, we define the concept of "dominant neurons." Specifically, we aim to assess the importance of each neuron for the forgetting client and its importance for all non-forgetting clients.

First, we select a list of sensitive neurons from all clients that belong to the same class as the forgetting data category and iterate through each neuron $N$. For each neuron $N$, we calculate its contribution to the forgetting client (Client 0), denoted as $S_{\text{forget}}$. Next, we search for the list of sensitive neurons that belong to the same class as this neuron $N$ in all non-forgetting clients (Client 1, Client 2, \dots), and find the maximum contribution of neuron $N$ in each non-forgetting client, denoted as $S_{\text{max\_other}}$. If $N$ does not appear in the list of any non-forgetting client, then $S_{\text{max\_other}} = 0$.

Then, we calculate the ratio $R$:
\begin{equation}
    R = \frac{S_{\text{max\_other}}}{S_{\text{forget}}}
\end{equation}

Based on the value of ratio $R$, we define the importance of a neuron as follows:
\begin{itemize}
    \item If $R$ is large, it indicates that the neuron is crucial for some non-forgetting clients and should not be modified arbitrarily.
    \item If $R$ is close to 1, it means the neuron has similar importance in both the forgetting and non-forgetting clients, and is referred to as a "shared neuron."
    \item If $R$ is small, it indicates that the contribution of the neuron to the forgetting client is significantly greater than its contribution to any non-forgetting client, and it can be considered a "dominant neuron."
\end{itemize}

Based on this theory, we propose the strategy of "ranking neurons by dominant score from low to high and selecting the first $n$ neurons," called "Rank-Based Selection." We will then edit the model by using the indices $(l, w)$ of the selected top $n$ neurons and set the corresponding weights of these neurons to zero.

\subsection{Experimental Analysis}

\begin{figure}[htbp]
  \centering
  \includegraphics[width=0.9\linewidth]{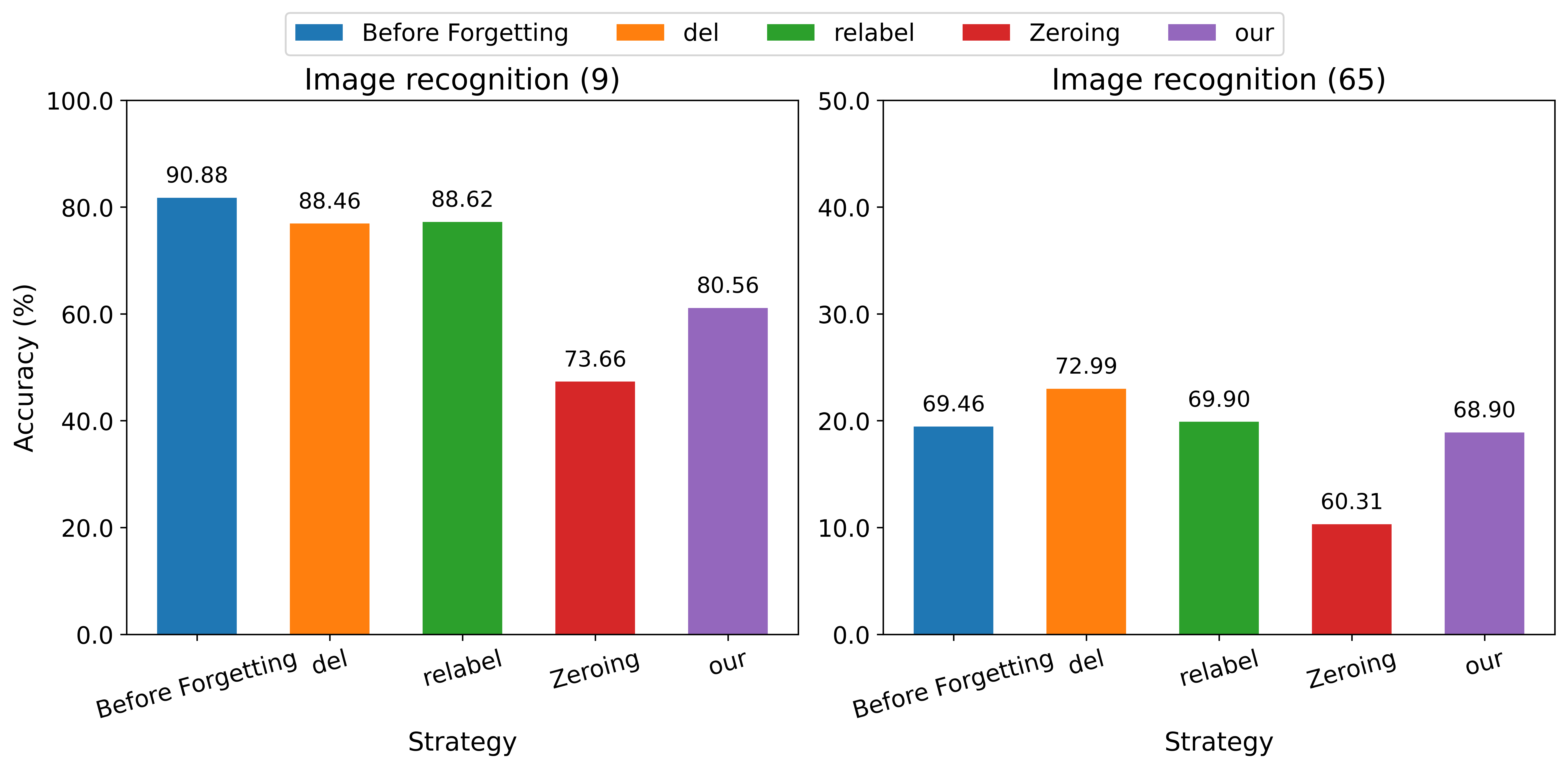}
  \caption{Performance Comparison of Different Unlearning Strategies on an Image Recognition Task}
  \label{fig4}
\end{figure}

After modifying the model, we selected more complex Image Recognition tasks (Image Recognition9 and Image Recognition65) for experimental analysis. The dataset was then partitioned using the Real-Noniid method, and the experiments were conducted following the principle of training fairness. Figure~\ref{fig4} presents the overall accuracy changes before and after unlearning. We can observe that the overall accuracy drop for the Delete-Retrain and Relabel-Poison techniques is quite limited after unlearning. Next is our proposed method, and finally, the zeroing-out approximate unlearning technique, which causes a significant decrease in overall accuracy.

To provide a more comprehensive analysis, Table~\ref{tab:Image_recognition9} and Table~\ref{tab:Image_recognition65} show the accuracy performance of the top three clients across different categories. This table details the accuracy on both the class targeted for unlearning (the forgotten class) and all remaining classes. This enables a more detailed observation of the effects of different unlearning methods across various categories. For the 65-category Image Recognition task, due to the large number of classes, we only present the accuracy of the top 10 categories for the first three clients to provide a clearer view of the results.

\begin{table}[htbp]
\centering
\renewcommand{\arraystretch}{0.9}
\footnotesize 
\setlength{\tabcolsep}{1pt}
\begin{tabular}{@{}ll rrrrr@{}}
\toprule
\textbf{Client} & \textbf{Label} & \textbf{Before} & \textbf{Delete} & \textbf{Relabel} & \textbf{Zeroing} & \textbf{Our} \\ 
\midrule
\multirow{9}{*}{Client 1} & label\_0($\downarrow$) & 94.33\% & 82.77\% & 83.67\% & \textbf{7.94\%} & 16.55\% \\
& label\_1 & 96.30\% & 96.54\% & 97.28\% & 92.35\% & 94.57\% \\
& label\_2 & 90.89\% & 90.65\% & 93.29\% & 79.86\% & 89.69\% \\
& label\_3 & 88.78\% & 89.28\% & 85.04\% & 89.28\% & \textbf{90.52\%} \\
& label\_4 & 91.14\% & 87.95\% & 92.50\% & 72.73\% & 87.27\% \\
& label\_5 & 95.95\% & 96.19\% & 96.67\% & 77.62\% & 95.48\% \\
& label\_6 & 95.58\% & 95.09\% & 95.82\% & 80.59\% & 93.12\% \\
& label\_7 & 97.06\% & 96.08\% & 97.79\% & 83.33\% & 95.10\% \\
& label\_8 & 96.22\% & 96.89\% & 96.00\% & 87.56\% & 95.78\% \\ 
\midrule
\multirow{9}{*}{Client 2} & label\_0 & 100.00\% & 77.78\% & 77.78\% & 5.56\% & \textbf{27.78\%} \\
& label\_1 & 88.89\% & 94.44\% & 100.00\% & 83.33\% & 88.89\% \\
& label\_2 & 88.46\% & 84.62\% & 84.62\% & 84.62\% & 88.46\% \\
& label\_3 & 91.30\% & 86.96\% & 91.30\% & 100.00\% & 86.96\% \\
& label\_4 & 86.36\% & 81.82\% & 81.82\% & 59.09\% & 86.36\% \\
& label\_5 & 90.91\% & 86.36\% & 90.91\% & 72.73\% & \textbf{90.91\%} \\
& label\_6 & 100.00\% & 95.24\% & 95.24\% & 90.48\% & \textbf{95.24\%} \\
& label\_7 & 100.00\% & 100.00\% & 100.00\% & 95.00\% & \textbf{100.00\%} \\
& label\_8 & 84.62\% & 92.31\% & 84.62\% & 76.92\% & 84.62\% \\ 
\midrule
\multirow{9}{*}{Client 3} & label\_0 & 94.05\% & 83.78\% & 84.86\% & 9.73\% & \textbf{20.54\%} \\
& label\_1 & 97.03\% & 97.03\% & 97.03\% & 91.09\% & 95.05\% \\
& label\_2 & 92.75\% & 91.19\% & 92.23\% & 76.68\% & 91.71\% \\
& label\_3 & 89.42\% & 88.46\% & 87.02\% & 87.50\% & \textbf{88.46\%} \\
& label\_4 & 92.31\% & 92.31\% & 91.72\% & 78.70\% & 89.35\% \\
& label\_5 & 97.79\% & 97.24\% & 97.79\% & 79.01\% & 97.24\% \\
& label\_6 & 95.63\% & 98.06\% & 97.57\% & 76.70\% & 96.12\% \\
& label\_7 & 97.21\% & 96.09\% & 97.21\% & 86.03\% & \textbf{97.21\%} \\
& label\_8 & 96.92\% & 98.46\% & 96.41\% & 92.31\% & 96.41\% \\ 
\bottomrule
\end{tabular}

\caption{Comparison of different unlearning methods on the Image Recognition(9) task across three clients. }
\label{tab:Image_recognition9}
\end{table}

From Table~\ref{tab:Image_recognition9}, we observe that while Delete-Retrain and Relabel-Poison result in only minor degradation in overall model performance, their forgetting efficacy remains limited. In contrast, the Zeroing route achieves stronger forgetting on the target client (e.g., class 0 accuracy on client1 drops from 94.33\% to 7.94\%), but introduces severe collateral effects. For example, class 0 accuracy on client2 and client3 drops to 5.56\% and 9.73\%, respectively, and class 4 on client2 decreases by 27.27\%.

In comparison, our method reduces class 0 accuracy on client1 to 16.55\%, achieving a comparable forgetting effect while substantially mitigating side effects. The accuracy of non-forgotten classes and clients remains largely unaffected. For instance, class 6 on client2 drops by only 4.76\%. These results demonstrate the effectiveness of our approach in balancing forgetting performance and cross-client stability.

\begin{table}[htbp]
\centering
\renewcommand{\arraystretch}{0.9}
\footnotesize 
\setlength{\tabcolsep}{1pt}
\begin{tabular}{@{}ll rrrrr@{}}
\toprule
\textbf{Client} & \textbf{Label} & \textbf{Before} & \textbf{Delete} & \textbf{Relabel} & \textbf{Zeroing} & \textbf{Our} \\ 
\midrule
\multirow{10}{*}{Client 1} & label\_0($\downarrow$) & 50.00\% & 50.00\% & 50.00\% & \textbf{0.00\%} & 50.00\% \\
& label\_1 & 100.00\% & 100.00\% & 100.00\% & 100.00\% & 100.00\% \\
& label\_2 & 50.00\% & 100.00\% & 100.00\% & 50.00\% & 50.00\% \\
& label\_3 & 100.00\% & 100.00\% & 100.00\% & 100.00\% & 100.00\% \\
& label\_4 & 50.00\% & 50.00\% & 0.00\% & 0.00\% & \textbf{50.00\%} \\
& label\_5 & 100.00\% & 100.00\% & 100.00\% & 100.00\% & 100.00\% \\
& label\_6 & 33.33\% & 33.33\% & 33.33\% & 33.33\% & 33.33\% \\
& label\_7 & 100.00\% & 100.00\% & 100.00\% & 0.00\% & 100.00\% \\
& label\_8 & 0.00\% & 0.00\% & 0.00\% & 0.00\% & 0.00\% \\
& label\_9 & 33.33\% & 33.33\% & 0.00\% & 0.00\% & 33.33\% \\ 
\midrule
\multirow{10}{*}{Client 2} & label\_0 & 53.85\% & 38.46\% & 53.85\% & 0.00\% & \textbf{53.85\%} \\
& label\_1 & 62.50\% & 62.50\% & 68.75\% & 56.25\% & 56.25\% \\
& label\_2 & 25.00\% & 50.00\% & 25.00\% & 12.50\% & 25.00\% \\
& label\_3 & 81.82\% & 90.91\% & 81.82\% & 72.73\% & 72.73\% \\
& label\_4 & 63.64\% & 81.82\% & 63.64\% & 54.55\% & 63.64\% \\
& label\_5 & 100.00\% & 100.00\% & 100.00\% & 100.00\% & 100.00\% \\
& label\_6 & 69.23\% & 38.46\% & 53.85\% & 53.85\% & \textbf{69.23\%} \\
& label\_7 & 87.50\% & 87.50\% & 87.50\% & 62.50\% & \textbf{87.50\%} \\
& label\_8 & 68.75\% & 81.25\% & 75.00\% & 62.50\% & 68.75\% \\
& label\_9 & 44.44\% & 33.33\% & 55.56\% & 44.44\% & 44.44\% \\ 
\midrule
\multirow{10}{*}{Client 3} & label\_0 & 76.74\% & 62.79\% & 69.77\% & 0.00\% & 55.81\% \\
& label\_1 & 73.17\% & 68.29\% & 63.41\% & 63.41\% & \textbf{70.73\%} \\
& label\_2 & 30.95\% & 35.71\% & 30.95\% & 19.05\% & 28.57\% \\
& label\_3 & 75.56\% & 88.89\% & 64.44\% & 53.33\% & 68.89\% \\
& label\_4 & 61.54\% & 69.23\% & 56.41\% & 43.59\% & 61.54\% \\
& label\_5 & 88.24\% & 88.24\% & 85.29\% & 85.29\% & \textbf{88.24\%} \\
& label\_6 & 65.79\% & 55.26\% & 65.79\% & 60.53\% & \textbf{63.16\%} \\
& label\_7 & 63.89\% & 86.11\% & 80.56\% & 47.22\% & 66.67\% \\
& label\_8 & 56.52\% & 60.87\% & 47.83\% & 41.30\% & 54.35\% \\
& label\_9 & 32.50\% & 32.50\% & 22.50\% & 17.50\% & \textbf{32.50\%} \\ 
\bottomrule
\end{tabular}

\caption{Comparison of different unlearning methods on the Image Recognition(65) task across three clients. }
\label{tab:Image_recognition65}
\end{table}

From Table~\ref{tab:Image_recognition65}, our method exhibits a consistent pattern in the more complex Image Recognition(65) task. While the Zeroing route causes severe degradation on non-forgotten data (e.g., a 33.33\% drop in class 9 on client1), indicating that our strategy effectively mitigates side effects on other data classes while preserving the unlearning effect.

\section{Conclusion}

We rethink the foundations of FU and show that two implicit assumptions, unfair global retraining and synthetic data partitions, have systematically inflated the reported effectiveness of FU mainstream technical routes. Building on this insight, we introduce \ours, an evaluation framework that mirrors practical deployment conditions, and demonstrate through extensive experiments that mainstream technical routes remain fragile in both fairness and forgetting quality.

Although our study charts an initial path toward fair and deployable FU, key challenges remain in balancing unlearning precision and minimizing cross-client side effects. We encourage future work to enhance fairness-aware unlearning toward robust real-world deployment.

\bibliography{aaai2026}

\setlength{\leftmargini}{20pt}
\makeatletter\def\@listi{\leftmargin\leftmargini \topsep .5em \parsep .5em \itemsep .5em}
\def\@listii{\leftmargin\leftmarginii \labelwidth\leftmarginii \advance\labelwidth-\labelsep \topsep .4em \parsep .4em \itemsep .4em}
\def\@listiii{\leftmargin\leftmarginiii \labelwidth\leftmarginiii \advance\labelwidth-\labelsep \topsep .4em \parsep .4em \itemsep .4em}\makeatother

\setcounter{secnumdepth}{0}
\renewcommand\thesubsection{\arabic{subsection}}
\renewcommand\labelenumi{\thesubsection.\arabic{enumi}}

\newcounter{checksubsection}
\newcounter{checkitem}[checksubsection]

\newcommand{\checksubsection}[1]{%
  \refstepcounter{checksubsection}%
  \paragraph{\arabic{checksubsection}. #1}%
  \setcounter{checkitem}{0}%
}

\newcommand{\checkitem}{%
  \refstepcounter{checkitem}%
  \item[\arabic{checksubsection}.\arabic{checkitem}.]%
}
\newcommand{\question}[2]{\normalcolor\checkitem #1 #2 \color{blue}}
\newcommand{\ifyespoints}[1]{\makebox[0pt][l]{\hspace{-15pt}\normalcolor #1}}

\end{document}